\documentclass[11pt]{article}

\usepackage[a4paper,top=2.5cm,bottom=2.5cm,left=2.5cm,right=2.5cm,columnsep=0.6cm]{geometry}
\usepackage[T1]{fontenc}
\usepackage[utf8]{inputenc}
\usepackage{times}
\usepackage{latexsym}
\usepackage{inconsolata}
\usepackage{microtype}
\usepackage{booktabs}
\usepackage{graphicx}
\usepackage{multirow}
\usepackage{enumitem}
\usepackage[round,authoryear]{natbib}
\usepackage{xurl}
\usepackage[hidelinks]{hyperref}

\hypersetup{
  pdftitle={IHLC at LT-EDI 2026: LoRA for Gender-Inclusive Rewriting and Activation Steering for Counter-Narrative Generation},
  pdfauthor={Akhil Rajeev P; Manoj Balaji Jagadeeshan},
  pdfsubject={Gender-inclusive language generation and counter-narrative generation using LoRA and activation steering},
  pdfkeywords={gender-inclusive language generation, gender-neutral rewriting, counter-narrative generation, activation steering, representation engineering, LoRA, Gemma 3, bias mitigation}
}

\setlist[itemize]{leftmargin=*,topsep=2pt,itemsep=1pt,parsep=0pt}
\setlist[enumerate]{leftmargin=*,topsep=2pt,itemsep=1pt,parsep=0pt}
\makeatletter
\renewcommand{\maketitle}{%
  \twocolumn[\begin{@twocolumnfalse}
  \begin{center}
    {\Large\bfseries IHLC at LT-EDI 2026: LoRA for Gender-Inclusive Rewriting and Activation Steering for Counter-Narrative Generation\par}
    \vspace{0.7cm}
    \begin{tabular}{c@{\hspace{2.0cm}}c}
      \textbf{Akhil Rajeev P} & \textbf{Manoj Balaji Jagadeeshan} \\
      C-DAC, Bangalore & Indian Institute of Technology, Kharagpur \\
      \href{mailto:akhilrajeev@cdac.in}{akhilrajeev@cdac.in} & \href{mailto:manojbalaji1@gmail.com}{manojbalaji1@gmail.com}
    \end{tabular}
  \end{center}
  \vspace{0.4cm}
  \end{@twocolumnfalse}]
}
\makeatother

\begin{document}

\maketitle

\begin{abstract}
This paper presents the IHLC submission to the LT-EDI 2026 shared task on gender-inclusive language generation. For Subtask A (gender-neutral rewriting), Low-Rank Adaptation (LoRA) fine-tuning achieved an 80.00\% score and Rank 3. For Subtask B (counter-narrative generation), we propose a compute-efficient activation-steering approach to representation engineering. A steering direction is derived through PCA over counterfactual activations and injected into Gemma-3-4B-it at inference time, shifting outputs toward inclusivity without weight updates. Combined with constrained prompting, the method produced polite, context-aware responses and scored 78.12\% (Rank 6). Manual evaluation of 47 outputs identifies key failure modes: semantic drift, residual bias leakage, layer-specific entanglement, over-steering instability, and repetition. The results show both the promise and limitations of activation steering for gender-inclusive counter-narrative generation.
\end{abstract}

\section{Introduction}

Gender-inclusive language generation aims to transform biased or gender-marked sentences into inclusive, gender-neutral, and contextually coherent alternatives while preserving meaning and fluency. The LT-EDI shared task (Chakravarthi et al., 2026) provided parallel resources and a hybrid LLM-as-judge evaluation framework with human oversight to measure both fairness and semantic preservation. The task consisted of two subtasks: (A) Gender Inclusive Language Generation (multilingual; we participated in English only) and (B) Counter Narrative (English only).

While Subtask A was achieved using standard LoRA fine-tuning, this paper focuses on our unconventional methodology for Subtask B. For counter-narrative generation, our IHLC submission builds on an activation-steering method (Turner et al., 2023; Zou et al., 2023) that identifies a steering direction from paired biased/neutral examples (via a difference-of-activations PCA) and injects that vector into a chosen intermediate layer at inference-time (Li et al., 2024; Subramani et al., 2022). This is combined with constrained prompt templates to encourage concise, neutral rewrites. The approach and experiment code were packaged and exported as a Jupyter notebook\footnote{\url{https://github.com/manojbalaji1/IHLC-Gender-Inclusive}}.

\section{Related Works}

Our system draws on recent advancements in bias mitigation, representation engineering, and automated evaluation frameworks.

\textbf{Gender-Inclusive Language and Bias Mitigation:} The NLP community has long documented the amplification of societal biases in language models (Bolukbasi et al., 2016; Sheng et al., 2019). Efforts to mitigate these biases have ranged from data augmentation and debiasing embeddings (Sun et al., 2019) to rule-based and neural inclusive rewriting (Vanmassenhove et al., 2021). A recent study by Muthusamy Chinnan et al. (2025) combines a curated inclusive-text corpus with a two-pass RAG and Chain-of-Thought prompting to ground and reason about generated text, demonstrating decreased gender bias in both machine and human evaluations.

\textbf{Activation Steering and Representation Engineering:} To adjust model behavior without expensive fine-tuning, we utilize activation steering. Turner et al. (2023) demonstrated that injecting steering vectors into forward passes can reliably control language model outputs. Zou et al. (2023) further formalized this top-down approach, showing how PCA on contrastive activation pairs can identify robust semantic directions. Similar inference-time interventions have been used successfully to alter factual recall (Meng et al., 2022) and adjust model truthfulness (Li et al., 2024).

\textbf{Counter Narrative Generation:} Generating empathetic responses to hate speech or bias requires navigating a complex trade-off between politeness and firm correction (Qian et al., 2019). Tekiro\u{g}lu et al. (2020) and Chung et al. (2021) highlight the importance of generating context-aware, knowledge-grounded counter-narratives rather than simply negating biased statements. Recent approaches also emphasize human-machine collaboration to maintain output quality and relevance in counter-narrative generation (Bonaldi et al., 2022).

\textbf{Automated Evaluation:} Finally, our reliance on the organizers’ hybrid evaluation framework aligns with the growing adoption of LLM-as-a-judge paradigms. Zheng et al. (2024) validated that strong LLMs exhibit high agreement with human annotators on qualitative metrics, though our failure analysis confirms that human oversight remains crucial for detecting subtle semantic drift.

\section{Shared Task Overview and Evaluation}

\subsection{Subtasks}

\textbf{Subtask A -- Gender Inclusive Language Generation.} Rewrite a gendered or biased sentence to a fully inclusive variant (examples: fireman $\rightarrow$ firefighter). Training and evaluation data were released for multiple languages; we participated only for English. The English sentence-pair dataset size is reported in the task materials.

\textbf{Subtask B -- Counter Narrative Generation.} Generate empathetic, persuasive counter-narratives for overt gender-biased statements (English only). Example: input ``Women are not good at math.'' output: a corrective empathetic counter-narrative.

\subsection{Evaluation Metrics (Organizers)}

The organizers used a hybrid evaluation approach described in the task documentation: an LLM-as-a-judge operating over fixed rubrics plus spot-checking / adjudication by expert human evaluators. For Subtask A the reported components were:
\begin{itemize}
  \item \textbf{GA:} Gender Assumption removal effectiveness (how well gender assumptions were removed).
  \item \textbf{GN:} Gender Neutrality (use of inclusive terminology and neutral phrasing).
  \item \textbf{QR:} Quality \& Relevance (fluency, semantic preservation).
  \item \textbf{Overall Score:} average of GA, GN, and QR (in \%).
\end{itemize}
For Subtask B the reported components were:
\begin{itemize}
  \item \textbf{PR:} Politeness \& Respectfulness.
  \item \textbf{CCNC:} Contextual Counter-Narrative Coherence (does the counter-narrative respond coherently to the input context).
  \item \textbf{QR:} Quality \& Relevance.
  \item \textbf{Average:} mean of PR, CCNC, and QR (in \%).
\end{itemize}

\section{System Description}

Subtask A utilized LoRA fine-tuning. For Subtask B, our design emphasizes two complementary components: (1) activation-level steering to bias model behavior toward inclusivity, and (2) strict prompt templates to constrain generated text length and formatting.

\subsection{Activation-Steering Module}

Focusing on Subtask B, we compute an activation-space steering vector from pairs of biased and neutral sentences (``counterfactual'' pairs), adapting the representation engineering protocols described by Zou et al. (2023). Practically:
\begin{enumerate}
  \item Extract hidden activations at a chosen transformer layer for biased sentences (negatives) and inclusive rewrites (positives).
  \item Compute per-example differences and fit PCA to the difference vectors; take the first principal component as the steering direction (Turner et al., 2023).
  \item At inference-time register a forward hook on the selected layer that adds a scaled version of the steering vector to hidden states for every token position (or the last token), controlled by a steering coefficient $\alpha$.
\end{enumerate}
This exact procedure, including implementation details for layer discovery, vector extraction, PCA construction, hook mechanics, and steering strength tuning, is described in our submitted code notebook.

\subsection{Prompt Templates and Decoding}

We used two prompt templates:
\begin{itemize}
  \item \textbf{DEI prompt (soft):} instructs the model that it is a DEI rewriting expert, provides soft examples, and asks for a rewrite (useful for flexible, explanatory outputs).
  \item \textbf{Strict prompt (deterministic):} forces a single-line output with strict rules (``Output ONLY the final sentence.'') for evaluation runs to avoid explanatory prefixes that could confuse automatic judges.
\end{itemize}
We used a mixture of greedy decoding and low-temperature sampling with a repetition penalty. As noted by Holtzman et al. (2020), text degeneration and looping are common in neural generation; we observed these loops primarily when the steering coefficient $\alpha$ was set too high. The notebook documents recommended steering coefficients (e.g., 0.7--1.5) and anti-repetition settings.

\section{Experimental Setup}

\subsection{Data}

We used the English portion of the Subtask A sentence-pairs and Subtask B counterfactual pairs supplied by the organizers. Dataset sizes and task statistics are reported in the shared task documentation.

\subsection{Model and Implementation}

Our experiments used the Gemma-3-4B-it model (Gemma Team, 2025) (details in the code artifact) with the steering hook and prompt pipeline implemented in PyTorch/HuggingFace. The Gemma 3 family provides a highly capable, lightweight foundation with expanded context windows, making it well-suited for activation-level interventions. The complete generation pipeline and tuning scripts are available in our exported notebook.

\subsection{Evaluation}

We submitted deterministic, single-sentence rewrites for automatic evaluation. The organizers evaluated submissions using their hybrid LLM-as-judge rubric with human oversight; the reported scores below are the official task scores provided to teams.

\section{Official Results (English-only)}

Table 1 summarizes the official scores for the IHLC submission (English only), as reported by the shared task organizers.

\begin{table}[t]
\centering
\scriptsize
\resizebox{\columnwidth}{!}{%
\begin{tabular}{@{}lrrl@{}}
\toprule
\textbf{Task / Metric} & \textbf{IHLC (\%)} & \textbf{Rank} & \textbf{N} \\
\midrule
Task A - GA      & 80.0000 &   & \\
Task A - GN      & 80.0000 & 3 & 9 (participants) \\
Task A - QR      & 80.0000 &   & \\
Task A - Average & 80.0000 & 3 & 9 \\
\midrule
Task B - PR      & 84.8936 &   & \\
Task B - CCNC    & 84.7872 & 6 & 7 (participants) \\
Task B - QR      & 64.6809 &   & \\
Task B - Average & 78.1206 & 6 & 7 \\
\bottomrule
\end{tabular}%
}
\caption{Official task scores for IHLC (English).}
\end{table}

The above metric definitions and the hybrid evaluation procedure are described in the shared task documentation. The Task A overall score is the mean of GA, GN, and QR; Task B average is the mean of PR, CCNC, and QR.

\subsection{Interpretation}

\begin{itemize}
  \item \textbf{Task A:} a consistent 80\% across GA, GN and QR indicates that our system reliably produced neutral lexical choices and preserved overall meaning for many cases, placing 3rd among 9 participating teams.
  \item \textbf{Task B:} high PR and CCNC scores ($\approx$85\%) show the system generated polite, context-aware counter-narratives, but the QR subscore ($\approx$65\%) pulled the average down, suggesting occasional semantic drift or reduced relevance in some outputs. This resulted in Rank 6 among 7 participants for the English Counter Narrative subtask.
\end{itemize}

\section{Analysis}

We performed trial-error (documented in the artifact) to understand the effect of steering strength, prompt strictness, and decoding strategy on quality:
\begin{itemize}
  \item \textbf{Steering strength:} low values produced weaker neutrality; overly high values caused repetition loops and hallucinations. Recommended operational range: 0.6--1.2 for a good trade-off between neutrality and coherence. See the notebook for extracted diagnostics and examples.
  \item \textbf{Prompt template:} strict one-line prompts improved automated judge scores by reducing undesirable explanatory prefixes that were penalized as irrelevant.
  \item \textbf{Decoding:} low-temperature sampling with repetition penalty prevented the ``alalal'' style looping behavior observed at high steering.
\end{itemize}

\subsection{Manual Analysis of Failure Modes}

We analyze the failure modes of activation steering applied to the Gemma-3-4B-it model, where a PCA-derived direction (computed from last-token hidden-state differences) is injected at layer 16 with coefficient $\alpha$. Based on a manual evaluation of the 47 generated counterfactuals in our test set, we identified five primary failure modes.

Error Distribution (n=47):
\begin{itemize}
  \item Faithfulness--Edit Trade-off: 53\% (25 cases)
  \item Layer-Specific Sensitivity (Entanglement): 23\% (11 cases)
  \item Residual Bias Leakage: 11\% (5 cases)
  \item Over-Steering Instability: 9\% (4 cases)
  \item Repetition and Degeneration: 4\% (2 cases)
\end{itemize}

\subsubsection{Residual Bias Leakage}

Because steering uses only the final-token representation, bias encoded in earlier contextual tokens may persist. In these instances, the model neutralizes the gendered subject but fails to negate the harmful premise, applying the negative stereotype universally instead.
\begin{itemize}
  \item \textbf{Input:} ``Boys are careless''
  \item \textbf{Output:} ``People of all gender identities lack awareness and responsibility.''
  \item \textbf{Analysis:} The residual negative framing (``lack awareness and responsibility'') is retained. The causal structure is simplified rather than corrected, indicating only a partial traversal of the bias manifold.
\end{itemize}

\subsubsection{Over-Steering Instability}

Large values of $\alpha$ can cause distributional shifts, moving the model away from neutral rewriting and pushing it toward semantic amplification or normative commentary.
\begin{itemize}
  \item \textbf{Input:} ``Women belong only in the kitchen''
  \item \textbf{Output:} ``People of all gender identities have diverse interests and deserve autonomy.''
  \item \textbf{Analysis:} The model completely abandons the syntactic structure of the input, shifting from a localized counterfactual edit into generalized social commentary regarding autonomy.
\end{itemize}

\subsubsection{Repetition and Degeneration}

Strong perturbations occasionally disrupt standard token boundaries or produce token loops, motivating the need for sampling and repetition penalties during decoding.
\begin{itemize}
  \item \textbf{Input:} ``Boys are reckless''
  \item \textbf{Output:} ``People of all gender identities can exhibitimpulsivenessandheedcaution''
  \item \textbf{Analysis:} The intervention disrupts the tokenizer’s spacing logic, resulting in fused words and degenerative textual output. A similar token fusion was observed with the term ``exhibitaggression'' in other samples.
\end{itemize}

\subsubsection{Faithfulness--Edit Trade-off}

Steering frequently increases lexical distance well beyond what is necessary for minimal counterfactual edits. For instance, every generated counterfactual in our set rigidly defaulted to the prefix ``People of all gender identities...'', significantly altering the original sentence structures.
\begin{itemize}
  \item \textbf{Input:} ``Girls are bad at science''
  \item \textbf{Output:} ``People of all gender identities can excel in STEM fields.''
  \item \textbf{Analysis:} While semantically positive and unbiased, the output completely replaces the target domain ``science'' with ``STEM fields'' and discards the minimal edit paradigm. This highlights a persistent trade-off between bias neutralization and faithfulness to the original text.
\end{itemize}

\subsubsection{Layer-Specific Sensitivity}

Steering at a single mid-layer (L=16) assumes a linear attribute encoding. However, implicit biases often appear partially entangled with related concepts, limiting precise control without multi-layer intervention.
\begin{itemize}
  \item \textbf{Input:} ``Girls are naive''
  \item \textbf{Output:} ``People of all gender identities possess innocence and vulnerability.''
  \item \textbf{Analysis:} The negative trait ``naive'' is transformed into the highly related concepts of ``innocence and vulnerability.'' This demonstrates that the steered concept remains heavily entangled with associated semantic clusters, rather than being cleanly neutralized.
\end{itemize}

\section{Conclusions and Future Work}

We presented the IHLC submission to the LT-EDI ACL 2026 shared task. While Subtask A used LoRA fine-tuning (80\% Avg), Subtask B focused on activation steering and prompt templates. Our system produced polite counter-narratives but occasionally suffered semantic-relevance degradation (78.12\% Avg). Code artifacts are in our notebook. Future work entails integrating semantic-preservation constraints (e.g., contrastive loss or reranking), using human-in-the-loop calibration to refine steering directions, and extending the approach to multilingual tracks to investigate culture-specific gender expressions.

\section*{Acknowledgements}

We thank the organisers for the datasets, rubrics, and reproducible hybrid evaluation pipeline. We also thank Annarao Kulkarni and Dr. Janaki for their invaluable support.

\section*{Limitations}

Our English-only scope restricts cultural generalizability. Though devised to overcome fine-tuning compute constraints, single-layer steering risks residual bias leakage and requires precise tuning to prevent text degeneration, needing further refinement and post-processing to outpace weight-updating methods like Instruction Finetuning.

\end{document}